\documentclass{svproc}
\usepackage{url}
\usepackage{graphicx}
\usepackage{times}
\usepackage{xcolor}
\usepackage{soul}
\usepackage[utf8]{inputenc}
\usepackage[small]{caption}
\usepackage{amssymb}
\usepackage{algorithm}
\usepackage[noend]{algpseudocode}
\usepackage{amsmath}

\begin{document}
\mainmatter              
\title{Heterogeneous Proxytypes Extended:\\Integrating Theory-like Representations and Mechanisms with Prototypes and Exemplars}
\titlerunning{Heterogeneous Proxytypes Extended}  
%
\author{Antonio Lieto}
\authorrunning{Antonio Lieto} 
%
\tocauthor{Antonio Lieto}
\institute{University of Turin, Dept. of Computer Science, Italy \\
ICAR-CNR (Palermo), Cognitive Robotics and Social Sensing Lab, Italy \\
\email{lieto@di.unito.it}\\
\texttt{Pre-print. Final Version: in Springer Advances in Intelligent Systems and Computing, BICA}\\
\texttt{Video: https://vimeo.com/297052905}
}

\maketitle              

\begin{abstract}

The paper introduces an extension of the proposal according to which conceptual representations in cognitive agents should be intended as \emph{heterogeneous proxytypes}. 
The main contribution of this paper is in that it details how to reconcile, under a heterogeneous representational perspective, different theories of typicality about conceptual representation and reasoning. In particular, it provides a novel theoretical hypothesis - as well as a novel categorization algorithm called DELTA - showing how to integrate the representational and reasoning assumptions of the theory-theory of concepts with the those ascribed to the prototype and exemplars-based theories.  

\keywords{heterogeneous proxytypes, knowledge representation, cognitive agents, cognitive architectures, declarative memory.}
\end{abstract}
%

\section{Introduction}

The proposal of characterizing the representational system of cognitive artificial agents by considering conceptual representations as \emph{heterogeneous proxytypes} was introduced in \cite{lieto2014computational}\footnote{The expression \emph{heterogeneous proxytypes} refers to both a theoretical and computational hypothesis combining the proxytype theory of concepts with the so called heterogeneity approach to concept representation. The Section 3 of this paper contains a brief discussion of the proposal.} and has been recently employed and successfully tested in systems like DUAL-PECCS \cite{lieto2016dual,lieto15ijcai,lieto2016towards}, later integrated with diverse cognitive architectures such as ACT-R \cite{anderson2004integrated}, CLARION \cite{sun06clarion}, SOAR \cite{laird2012soar} and Vector-LIDA \cite{snaider2014vector}.  The main contribution of this work is in that it offers a proposal to reconcile, under a heterogeneous representational perspective, not only prototype and exemplars based representations and reasoning procedures, but also the representational and reasoning assumptions ascribed to the so called theory-theory of concepts \cite{murphy2002big}. 
In doing so, the paper proposes a novel  categorization algorithm, called \emph{DELTA} (i.e. unifie\textbf{D}  Cat\textbf{E}gorization a\textbf{L}gorithm for he\textbf{T}erogeneous represent\textbf{A}tions) able to unify and integrate, in a cognitively oriented perspective, all the common-sense categorization mechanisms available in the cognitive science literature. The rest of the paper is organized as follows: the Section 2 provides an overview of the main representational paradigms proposed by the Cognitive Science and the Cognitive Modelling communities. Section 3, briefly synthesize the representational framework intending concepts as \emph{heterogeneous proxytypes} by showing how such theoretical proposal has been actually implemented and successfully tested in the DUAL-PECCS system. Section 4, proposes a more close analysis of the findings of the theory-theory of concepts, while, Section 5, proposes a novel and extended categorization algorithm integrating the theory-theory representational and reasoning mechanisms with those involving both exemplars and prototypes.


\section{Prototypes, Exemplars, Theories and Proxytypes}\label{prototypes_exemplars_proxytypes}
In the Cognitive Science literature, different theories about the nature of concepts have been proposed. 
According to the so called classical theory, concepts can be simply defined in terms of sets of necessary and sufficient conditions. Such theory was dominant until the mid '$70$s of the last Century, when Rosch's experimental results demonstrated the inadequacy of such a theory for ordinary --or common-sense -- concepts ~\cite{rosch75cognitive}. Rosch's results suggested, on the other hand, that ordinary concepts are characterized and organized in our mind in terms of \emph{prototypes}. %
Since then, different theories of concepts have been proposed to explain different representational and reasoning aspects concerning the problem of typicality: the prototype theory, the exemplars theory and the theory-theory.
%
According to the \emph{prototype} view, knowledge about categories is stored in terms of prototypes, i.e., in terms of some  representation of the ``best'' instance of the category. In this view, the concept \textsl{bird} should coincide with a representation of a typical bird (e.g., a robin). In the simpler versions of this approach, prototypes are represented as (possibly weighted) lists of typical features. %
According to the \emph{exemplar} view, a given category is mentally represented as set of specific exemplars explicitly stored in memory: the mental representation of the concept \textsl{bird} is a set containing the representation of (some of) the birds we encountered during our past experience. %
Another well known typicality-based theory of concepts is the so called the \emph{theory-theory} \cite{murphy2002big}. Such approach adopts some form of holistic point of view about concepts. According to some versions of the theory-theories, concepts are analogous to theoretical terms in a scientific theory. For example, the concept \textsl{cat} is individuated by the role it plays in our mental theory of zoology. In other versions of the approach, concepts themselves are identified with micro-theories of some sort. For example, the concept \textsl{cat} should be identified with a mentally represented microtheory about cats.

Although these approaches have been largely considered as competing ones (since they propose different models and predictions about how we organize and reason on conceptual information), they turned out to be not mutually exclusive~\cite{malt1989line}. Rather, they seem to succeed in explaining different classes of cognitive \emph{phenomena}, 
such as the fact that human subjects use different representations to categorize concepts. In particular, it seems that we can use  - in different situations - exemplars, prototypes or theories \cite{smith1998prototypes,murphy2002big,keil1989concepts}.  
Such experimental evidences led to the development of the so called ``heterogeneous hypothesis'' about the nature of concepts: this approach assumes that concepts do not constitute a unitary phenomenon, and hypothesizes that different types of conceptual representations may co-exist: prototypes, exemplars, theory-like or classical representations ~\cite{machery2009doing}. 
All such representations, in this view, constitute different \emph{bodies of knowledge} and contain different types of information associated to the the same conceptual entity. 
Furthermore, each body of conceptual knowledge is assumed to be featured by specific processes in which such representations are involved (e.g., in cognitive tasks like recognition, learning, categorization, \emph{etc.}). In particular prototypes, exemplars and theory-like default representations are associated with the possibility of dealing with non-monotonic strategies of reasoning and categorization, while the classical representations (i.e. that ones based on necessary and/or sufficient conditions) are associated with standard deductive mechanism of reasoning \footnote{In order to explain the different categorization strategies associated to different kinds of representations, let us consider the following examples: if we have to categorize a stimulus with the following features: ``it has fur, woofs and wags its tail'', the result of a \emph{prototype-based categorization} would be \textsl{dog}, since these  cues are associated to the prototype of \textsl{dog}. Prototype-based reasoning, however, is not the only type of reasoning based on typicality. In fact, if an exemplar corresponding to the stimulus being categorized is available, too, it is acknowledged that humans use to classify it by evaluating its similarity w.r.t. the exemplar, rather than w.r.t. the prototype associated to the underlying concepts~\cite{frixione2013representing}. For example, a penguin is rather dissimilar from the prototype of \textsl{bird}. However, if we already know an exemplar of penguin, and if we know that it is an instance of \textsl{bird}, it is easier to classify a new penguin as a \textsl{bird} w.r.t. a categorization process based on the similarity with the prototype of that category. This type of common-sense categorization is known in literature as \emph{exemplars-based categorization}. An example of theory-like common sense reasoning is when we typically associate to a light switch the learned rule that if we turn it ``on" then the light will be provided (this is a non-monotonic inference with a defeasible conclusion). Finally, the classical representations (i.e. those based on necessary and/or sufficient conditions) are associated with standard deductive mechanism of reasoning. An example of standard deductive reasoning is the categorization as \textsl{triangle} of a stimulus described by the features: ``it is a polygon, it has three corners and three sides''. Such cues, in fact, are necessary and sufficient for the definition of the concept of triangle. All these representations, and the corresponding reasoning mechanisms, are assumed to be potentially co-existing according to the heterogeneity approach.}.


In recent years an alternative theory of concepts has been proposed: the \emph{proxytype theory}. It postulates a biological localization and interaction between different brain areas for dealing with conceptual structures. Such localization have a direct counterpart in the well known distinction between \emph{long term} and \emph{working memory}~\cite{prinz2002furnishing}. In addition, such characterization is particularly interesting for the explanation of phenomena such as, for example, the activation (and the retrieval) of conceptual information. In this setting, concepts are seen as \emph{proxytypes}. 
%
A \emph{proxytype} is any element of a complex representational network \emph{stored in long-term memory} corresponding to a particular \emph{category} that can be tokenized in working memory to `go proxy' for that category~\cite{prinz2002furnishing}.
%
\noindent
In other terms, the proxytype theory, inspired by the work of Barsalou~\cite{barsalou1999perceptual}, considers concepts as \emph{temporary constructs} of a given category, activated (tokenized) in working memory as a result of conceptual processing activities, such as concept identification, recognition and retrieval. 
\section{Heterogeneous Proxytypes}
In the original formulation of the proxytypes theory, however, proxytypes have been depicted as monolithic conceptual structures, primarily intended as prototypes~\cite{de2005prinz}. A revised view of this approach has been recently proposed, hypothesizing the availability of a wider range of representation types than just prototypes~\cite{lieto2014computational}. They correspond to the kinds of representations hypothesized by the above mentioned heterogeneous approach to concepts. 
In this sense, proxytypes are assumed to be heterogeneous in nature (i.e., they are assumed to be composed by heterogeneous networks of conceptual representations and not only by a monolithic one)\footnote{The heterogeneity assumption has been recently pointed out as one of the problems to face in order to address the problems affecting the knowledge level in cognitive systems and architecture \cite{lieto2017knowledge,AAAIStandardModel}.}.  

In this renewed formulation, heterogeneous representations (such as \emph{prototypes}, \emph{exemplars}, \emph{theory-like} structures, \emph{etc.}) for each conceptual category are assumed to be \emph{stored in long-term memory}. They can be activated and accessed by resorting to different categorization strategies. In this view, each representation has its associated accessing procedures. In the following, I will briefly present how such theoretical hypothesis has been implemented in the DUAL-PECCS categorization system, and I will use the latter system as a computational referent for showing how the proposals presented in this paper can extend both the system itself and, more importantly, its underlying theoretical framework.

%
\subsection{Heterogeneous Proxytypes in DUAL-PECCS}
\begin{figure}[!t]
	\centering	\includegraphics[width=1.03\textwidth]{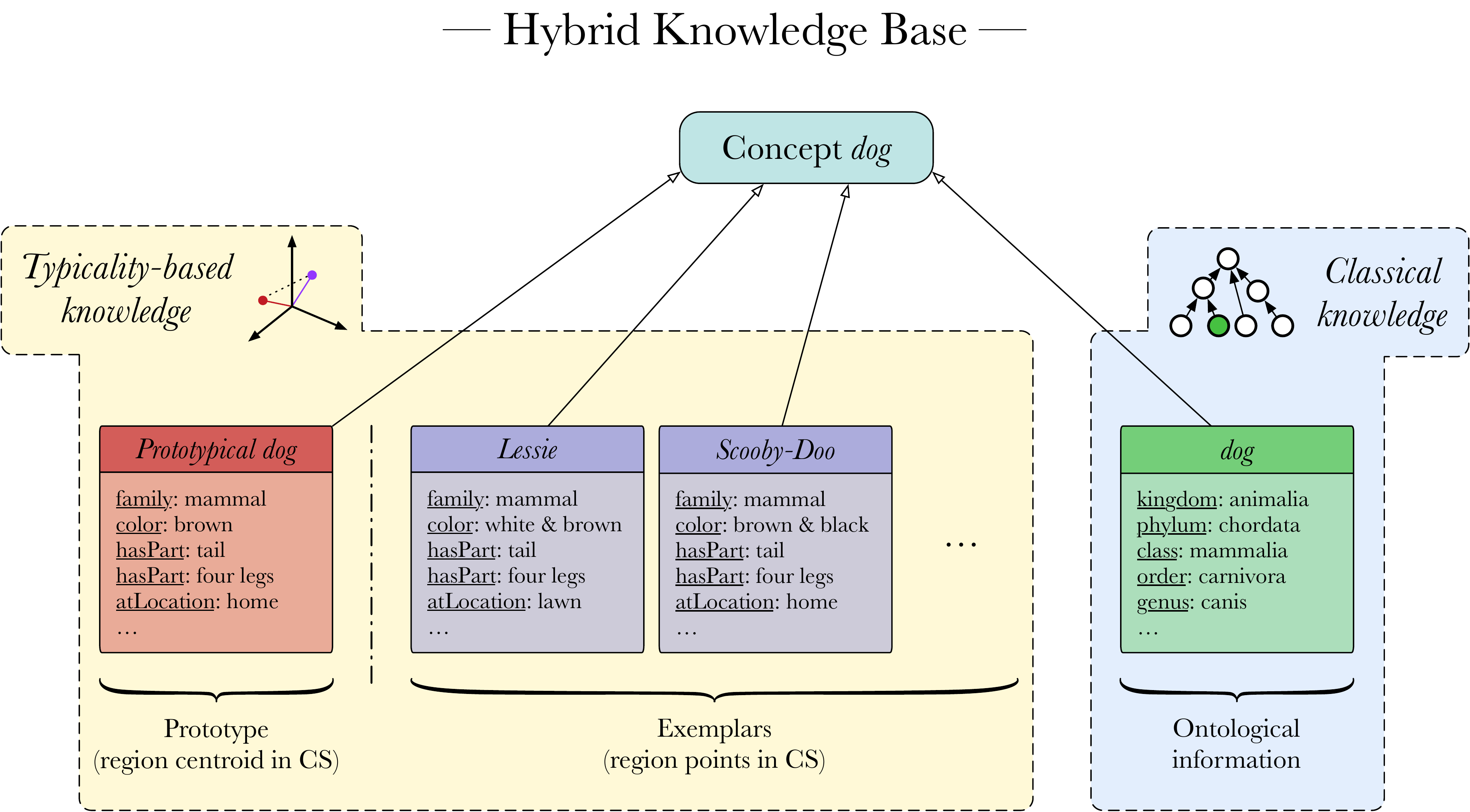}
		\caption{Heterogeneous representational architecture for the concept DOG in DUAL-PECCS.}
	\label{fig:dual}
\end{figure}

\noindent DUAL-PECCS \cite{lieto15ijcai,lieto2016dual}, is a cognitive categorization system explicitly designed and implemented under the heterogeneous proxytypes assumption\footnote{The characterization in terms of ``heterogeneous proxytypes'', among the other things, enables the system to deal with the problem of the ``contextual activation'' of a given information based on the external stimulus being considered. In particular (by following the idea that, when we categorize a stimulus, we do not activate the whole network of knowledge related to its assigned category but, conversely, we only activate the knowledge that is ``contextually relevant'' in its respect), DUAL-PECCS \emph{proxyfyes} only the type of representation that minimizes the distance w.r.t. the percept (see~\cite{lieto2014computational} for further details).} 
for both the representational level (that is: it is equipped with a hybrid knowledge base composed of heterogeneous representations, each endowed with specific reasoning mechanisms) and for the `proxyfication' mechanisms (i.e.: the set of procedures implementing the tokenization of the different representations in working memory).
The heterogeneous conceptual architecture of DUAL PECCS includes prototypes, exemplars and classical representations. All these different bodies of knowledge point to the same conceptual entity (the anchoring for these different types of representations is obtained via the Wordnet, see again \cite{lieto2016dual}). An example of the heterogeneous conceptual architecture of DUAL PECCS is provided in the Figure \ref{fig:dual}. Such figure  shows how it is represented the concept \textsl{dog}. 
In this case, the prototypical representation grasps information such as that dogs are usually conceptualized as domestic animals, with typically four legs, a tail \textit{etc.}; the exemplar-based representations grasp information on individuals. For example, in Fig.~\ref{fig:dual} it is represented the individual of \textsl{Lessie}, which is a particular exemplar of \textsl{dog} with white and brown fur and with a less domestic attitude w.r.t. the prototypical dog (e.g. its typical location is lawn). Within the system, both the exemplar and prototype-based representations make use of non classical (or typical) information and are represented by using the framework of the conceptual spaces \cite{gardenfors2004conceptual,lieto16conceptual}: a particular type of vector space model adopting standard similarity metrics to determine the distance between instances and concepts within the space. The representation of classical information (e.g. the fact that $\textsl{Dog} \sqsubseteq \textsl{Animal}$, that is to say that ``$Dogs$ are also $Animals$'') is, on the other hand, demanded to standard ontological formalisms. In the current version of the system the classical knowledge component is grounded in the OpenCyc ontology \cite{lenat1985cyc}.

By assuming the heterogeneous hypothesis, in DUAL-PECCS, different kinds of reasoning strategies are associated to these different bodies of knowledge. In particular, the system combines non-monotonic common-sense reasoning (associated to the \emph{prototypical} and \emph{exemplars-based}, conceptual spaces, representations) and standard monotonic categorization procedures (associated to the classical, ontological, body of knowledge). These different types of reasoning are harmonized according to the theoretical tenets coming from the dual process theories of reasoning and rationality~\cite{evans2009two,kahneman2011thinking}. 

As emerges from the figure \ref{fig:dual}, a missing part of the current conceptual architecture in the DUAL-PECCS system (and in its underlying theoretical hypothesis) concerns the representation of the default knowledge in terms of theory-like representational structures (while it already integrates classical, prototypical and exemplars based knowledge reprentation and processing mechanisms). In the next section we will show how Theory-like representations can be considered \emph{dual} in nature (at least from a formal point of view) and therefore may deserve a dual treatment also form a computational point of view.

%
%
%
%

%


\section{The Duality of Theory-Like Representations}

As mentioned in Section 2, Theory-theory approaches \cite{murphy1985role,murphy2002big} assume that concepts consists
of more or less complex mental structures representing (among other things) causal
and explanatory relations between them (including folk psychology connections). During the 80's, these approaches stemmed from a critique to the formerly dominant theory of concepts as prototypes. Consider, for example, the famous \cite{keil1989concepts} transformation experiments, in which subjects were asked to make categorization judgments about the biological membership of an animal that had undergone unusual transformations. In such experiment, Keil showed that people relies on theory-like representation (instead of prototypes) in order to execute their categorization task. 
In particular, it was shown that the type of knowledge retrieved by the subjects to solve these tasks belongs to their “default common-sense theory”  associated to a given concept. 

The idea that for most of our categories, our default knowledge includes a common-sense theory of that category  (and that theory-like default bodies of knowledge are associated with a distinct kind of categorization process) is, however, only one of the available interpretations about the theory-like representational structures \cite{machery2009doing}. Another kind of interpretation, in fact, assumes that theory-like structures do not constitute our typical default knowledge but that, on the other hand, they are constitutive of our classical background knowledge \cite{blanchard2010default}.
 In order to better explain this difference, and thus the \emph{duality} of the theory-like representations, let us consider the case of DUAL-PECCS.  
 As mentioned above, the current version of the system does not allow to represent the type of theory-like default knowledge belonging to the typical conceptual component of the architecture (see footnote 1 for an example of the non-monotonic reasoning that could be enabled by this kind of knowledge). On the other hand, it  allows to represent (in terms of IF-THEN rules enabling monotonic inferences), the kind of theory-like knowledge structures which are compliant with the ontological semantics of the classical conceptual component. 
In other words: only certain types of theories, i.e.  causal theories, belonging to the background knowledge of a cognitive agents, are currently covered by the integration of the current state of the art ontology languages and rules \cite{frixione14towards} in the DUAL-PECCS system. However, as already pointed out before, common sense knowledge is mostly characterized in terms of “theories” which are based on
 arbitrary, i.e. experience-based, rules. Therefore, in order to represent, within an artificial
 system, more realistic (from a cognitive standpoint) “theories”, i.e. including  common-sense default theories as intended in the theory-theory
approaches, there is the need of going beyond classical logic rules. 
Recently, graphical models (in particular Bayesian networks) have been proposed as a computational framework able to represent \cite{danks2007theory,danks2004psychological} knowledge networks of theory-like common-sense default representation. The integration of such framework within the DUAL-PECCS system represent a current and future area of development, not yet concluded. In the remaining of this paper, such integration will not be discussed and, in a certain sense, will be taken for granted. I shall focus, instead, on the presentation of a novel unifying categorization algorithm - named DELTA - able to harmonize all the different types of typicality-based representations and reasoning mechanisms associated with the common-sense knowledge: exemplars, prototypes and default theory-like representations. I will leave aside the discussion concerning the integration of such common-sense categorization mechanisms with those concerning the classical monotonic ones. As above mentioned, in fact, such integration is already provided in DUAL-PECCS \cite{lieto2016dual,lieto15ijcai} and is tackled by recurring to the dual process theory of reasoning (i.e.: the non monotonic reasoning results of the heterogeneous common-sense conceptual components are then checked and integrated with the monotonic reasoning strategies executed in the classical conceptual component). Therefore, the underlying \emph{heterogeneous-proxytypes} assumption, integrated with the dual process theory of reasoning, has been already proven to be effective to harmonize non monotonic and monotonic categorization strategies associated to heterogeneous body of knowledge. I will not report here the details of such harmonization procedure because it already documented elsewhere \cite{lieto2016dual}. I will focus, instead, on the harmonization procedures concerning the non monotonic categorization processes of the typical conceptual components.

\section{A Unified Categorization Algorithm for Exemplars, Prototypes and Theory-Like Representations}

In the following, I propose a novel categorization algorithm that, given a certain stimulus $d$, must select the most appropriate typicality-based representation available in the declarative memory of a cognitive agent (i.e. a prototype, an exemplar or a theory-like structure). According to what introduced in the previous sections, such declarative memory is assumed to rely on the \emph{heterogeneous proxytypes} hypothesis. 


%
The implemented procedure works as follows: when the input \emph{stimulus} is similar enough to an exemplar representation (a threshold has been fixed to these ends), the corresponding exemplar of a given category is retrieved. Otherwise, the prototypical representations are also scanned and the representation (prototype or exemplar) that is closest to the input is returned. By following a preference that has been experimentally observed in human cognition~\cite{medin1978context}, this algorithm favors the results of the exemplars-based categorization if the knowledge-base stores any exemplars similar to the input being categorized. 
As an additional constraint, I have hypothesized a mechanism in which theory-like structures of default knowledge can also override the categorization based on prototypes. Such mechanism has been devised based on the fact that theory-theorists have shown that, in some categorical judgments tasks (e.g. assessing the situation where a dog is made to look like a raccoon), categorization is driven by the possession of a rudimentary biological theory and by theory-like representations \cite{atran2008native}. In other words: being a dog isn't just a matter of looking like a dog. It seems, in fact, that it is more important to have a network of appropriate hidden properties of dogs: the dog ``essence''\cite{atran2008native}.
In the proposed algorithm, I have taken into account this element by hypothesizing: i) to measure the similarity between the theory-like representation of the first retrieved prototype with the stimulus $d$ \footnote{Since all the different bodies of knowledge are assumed to be co-referring representational structure pointing to the same conceptual entity, it is possible to recover the theory-like representation associated, for example, to a given prototypical or exemplar based representation.} and ii) to compare the obtained result with a \emph{Conceptual Coherence Threshold} that should measure how much the 
considered stimulus $d$ shares, i.e. is \emph{conceptually coherent}, with
the corresponding theory-like
representation of the retrieved
prototype. The analysis of the conceptual coherence can be solved as a constraint satisfaction problem as shown in \cite{thagard1998coherence}.


In this setting, if the distance between the stimulus $d$ and theory-like representation of the originally retrieved prototype is above the considered threshold, it means that the retrieved prototype is assumed to be representative enough of the common-sense ``essence' of $d$ (i.e it is ``coherent enough'). In this case, the prototypical answer is maintained, otherwise it is overridden by the theory-like representation which is closer to $d$.   

Let us assume, for example, that the stimulus to categorize is represented by an atypical Golden Zebra (which is almost totally white) and that in our agent's long-term memory there is no exemplar similar enough to this entity. This means that there will be no exemplar-based representation selected by our algorithm, and that the most similar representation to  $d$ will be searched among the prototypical representations in the agent knowledge base. Now: if we assume that the retrieved prototype is a typical white horse, we could discard such representation by simply relying on some additional information coming from the comparison of the stimulus $d$ (e.g. the fact that lives in the Savannah, etc.) with the default and common-sense theory associated to a horse (i.e. the category associated to the original prototypical choice). In this case the categorical assignment to the class Golden Zebra would be obtained by exploiting theory-like representational networks. 
A synthetic representation of the proposed procedure is presented in the Algorithm 1. 






\begin{figure}[!t]
	\centering	\includegraphics[width=1.05\textwidth]{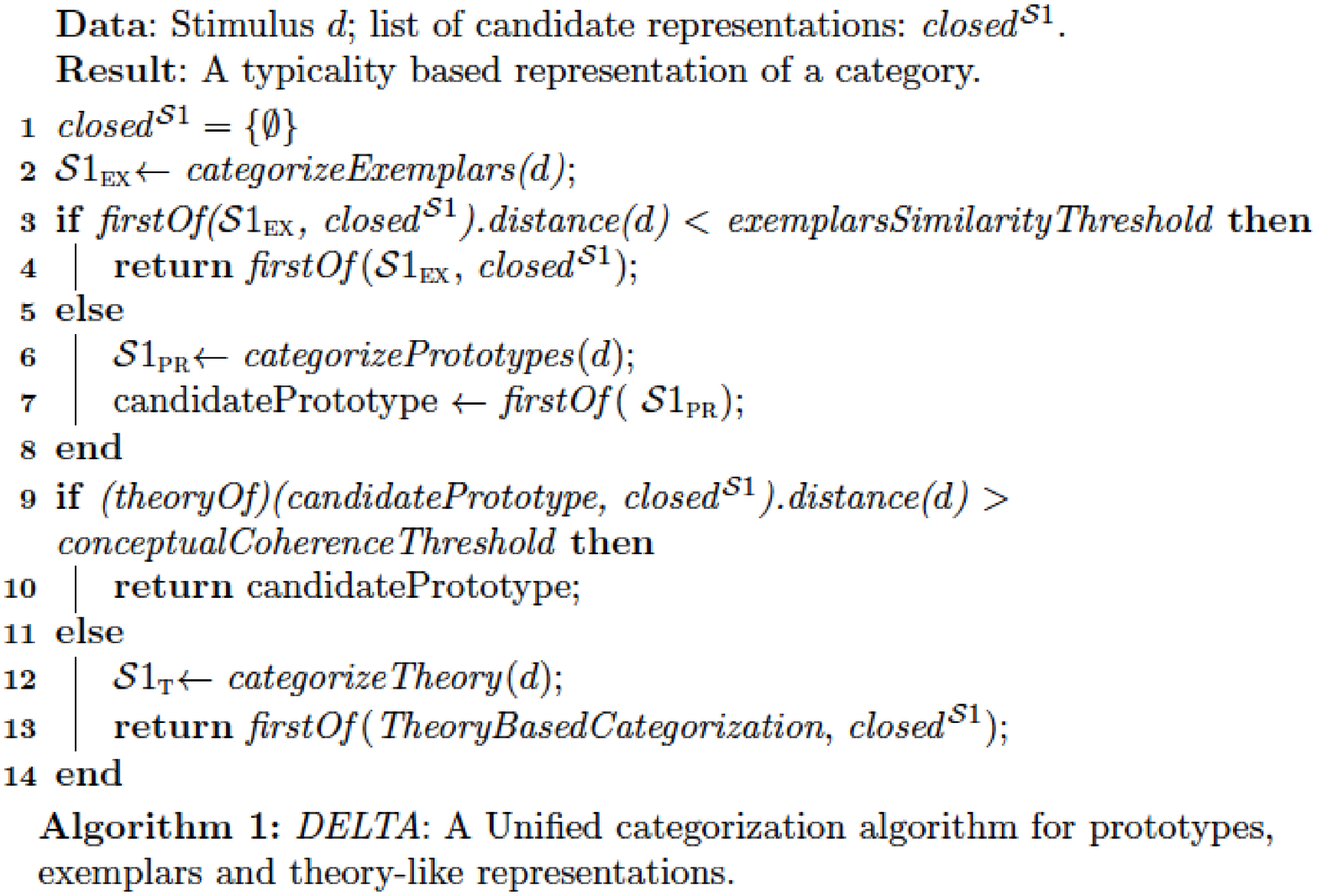}
	\label{fig:algo}
\end{figure}

\section{Conclusions and Future Work}
In this paper I have proposed a categorization algorithm able to unify all the common-sense categorization strategies proposed in the cognitive science literature: exemplars, prototypes and theory-like common-sense knowledge structures. To the best of my knowledge, this proposal represents the first attempt of providing a unifying categorization strategy by assuming a heterogeneous representational hypothesis. 
In particular, the proposed algorithm relies and extends both the representational and the reasoning framework considering concepts as heterogeneous proxytypes \cite{lieto2014computational}.
The current theoretical proposal needs to be tested on the empirical ground in order to show both its feasibility with psychological data and its efficacy in the area of artificial cognitive systems. Also, there are additional elements, only sketched in this paper, requiring a more precise characterization. For example: the design of a method to calculate which is the most appropriate theory-like representations to select (line 12, Algorithm 1). On this point, however, it is worth-noticing that, since the most promising computational candidates for representing the theory-like body of knowledge are graphical models and probabilistic semantic networks, it seems plausible to imagine that such calculation could be performed with standard heuristics search on graph structures. Similarly, the individuation and the construction of a plausible \emph{Conceptual Coherence Threshold} represents an issue that should be faced and solved empirically.\\


\textbf{Acknowledgements}
\\ 
The topics presented in this paper have been discussed in these years with a number people in international conferences, symposia, panels and workshops. I thank all them for the received comments. In particular, I am indebted to Marcello Frixione, Leonardo Lesmo, Paul Thagard, David Danks, Ismo Koponen and Christian Lebiere for their feedback and suggestions. I also thank Valentina Rho for her comments on an earlier version of this paper. 

\bibliographystyle{splncs03}
\bibliography{bibliography}








\end{document}